\documentclass{article}
\usepackage{times}
\usepackage{graphicx} 
\usepackage{subfigure} 
\usepackage{microtype}
\usepackage[sort&compress]{natbib}
\usepackage{algorithm}
\usepackage{algorithmic}
\usepackage{amsmath}
\usepackage{hyperref} 
\usepackage{flushend}
\usepackage{booktabs}

\newcommand{\abr}{CW-RNN}
\newcommand{\name}{Clockwork Recurrent Neural Network}

\let\oldtheequation\theequation
\makeatletter
\def\tagform@#1{\maketag@@@{\ignorespaces#1\unskip\@@italiccorr}}
\renewcommand{\theequation}{(\oldtheequation)}
\makeatother 

\usepackage[accepted]{icml2014ed}

\icmltitlerunning{A Clockwork RNN}

\begin{document} 

\twocolumn[
\icmltitle{A Clockwork RNN}

\icmlauthor{Jan Koutn\'{i}k}{hkou@idsia.ch}
\icmlauthor{Klaus Greff}{klaus@idsia.ch}
\icmlauthor{Faustino Gomez}{tino@idsia.ch}
\icmlauthor{J\"{u}rgen Schmidhuber}{juergen@idsia.ch}
\icmladdress{IDSIA, USI\&SUPSI,
            Manno-Lugano, CH-6928, Switzerland}

\icmlkeywords{recurrent neural network, sequence learning, prediction, speech classification}

\vskip 0.3in
]

\begin{abstract}

Sequence prediction and classification are ubiquitous and challenging
problems in machine learning that can require identifying complex
dependencies between temporally distant inputs. Recurrent Neural
Networks (RNNs) have the ability, in theory, to cope with these
temporal dependencies by virtue of the short-term memory implemented
by their recurrent (feedback) connections. However, in practice they
are difficult to train successfully when the long-term memory is
required.
This paper introduces a simple, yet powerful modification to the
standard RNN architecture, the {\it Clockwork RNN} (\abr), in which
the hidden layer is partitioned into separate modules, each processing
inputs at its own temporal granularity, making computations only at
its prescribed clock rate.
Rather than making the standard RNN models more complex, \abr{}
reduces the number of RNN parameters, improves the performance
significantly in the tasks tested, and speeds up the network evaluation.
The network is demonstrated in preliminary experiments involving two
tasks: audio signal generation and TIMIT spoken word classification,
where it outperforms both RNN and LSTM networks.
\end{abstract}

\section{Introduction}  %%%%%%%%%%%%%%%%%%%%%%%%%%%%%%%%%%%%%%%%%%%%
\label{sec:intro}
Recurrent Neural Networks
(RNNs;~\citealp{RobinsonFallside:87tr,Werbos:88gasmarket,Williams:89})
are a class of connectionist models that possess internal state or
{\em short term memory} due to recurrent feed-back connections, that
make them suitable for dealing with sequential problems, such as
speech classification, prediction and generation.

Standard RNNs trained with stochastic gradient descent have difficulty
learning long-term dependencies (i.e. spanning more that 10
time-steps) encoded in the input sequences due to the {\it vanishing
gradient}~\cite{Hochreiter:91,Hochreiter:01book}. 
The problem has been addressed for example by using a specialized neuron
structure, or cell, in Long Short-Term Memory (LSTM)
networks~\cite{Hochreiter:97lstm} that maintains constant backward flow
in the error signal; second-order optimization
methods~\cite{ICML2011Martens_532} preserve the gradient by estimating
its curvature; or using informed random
initialization~\cite{SutskeverMartensDahlHinton_icml2013} which allows
for training the networks with momentum and stochastic gradient
descent only.

This paper presents a novel modification to the simple RNN (SRN;
\citealp{elman:88}) architecture and, {\it mutatis mutandis},
an associated error back-propagation through time
\cite{rumelhart:86,Werbos:88gasmarket,Williams:89} training algorithm,
that show superior performance in the generation and classification of
sequences that contain long-term dependencies.  Here, the long-term
dependency problem is solved by having different parts (modules) of
the RNN hidden layer running at different clock speeds, timing their
computation with different, discrete clock periods, hence the name \name{}
(\abr). {}\abr{} train and evaluate faster since not all modules are
executed at every time step, and have a smaller number of weights
compared to SRNs, because slower modules are not connected to faster
ones.

\abr s were tested on two supervised learning tasks: sequence
generation where a target audio signal must be output by a network
using no input; and spoken word classification using the TIMIT
dataset. In these preliminary experiments, \abr{} outperformed both
SRN and LSTM with the same number of weights by a significant
margin. The next section provides an overview of the related work,
\autoref{sec:method} describes the \abr{} architecture in detail and
\autoref{sec:disc} discusses the results of experiments in
\autoref{sec:exp} and future potential of \name s.

\begin{figure*}[ht]
\centering
\includegraphics[width=\textwidth]{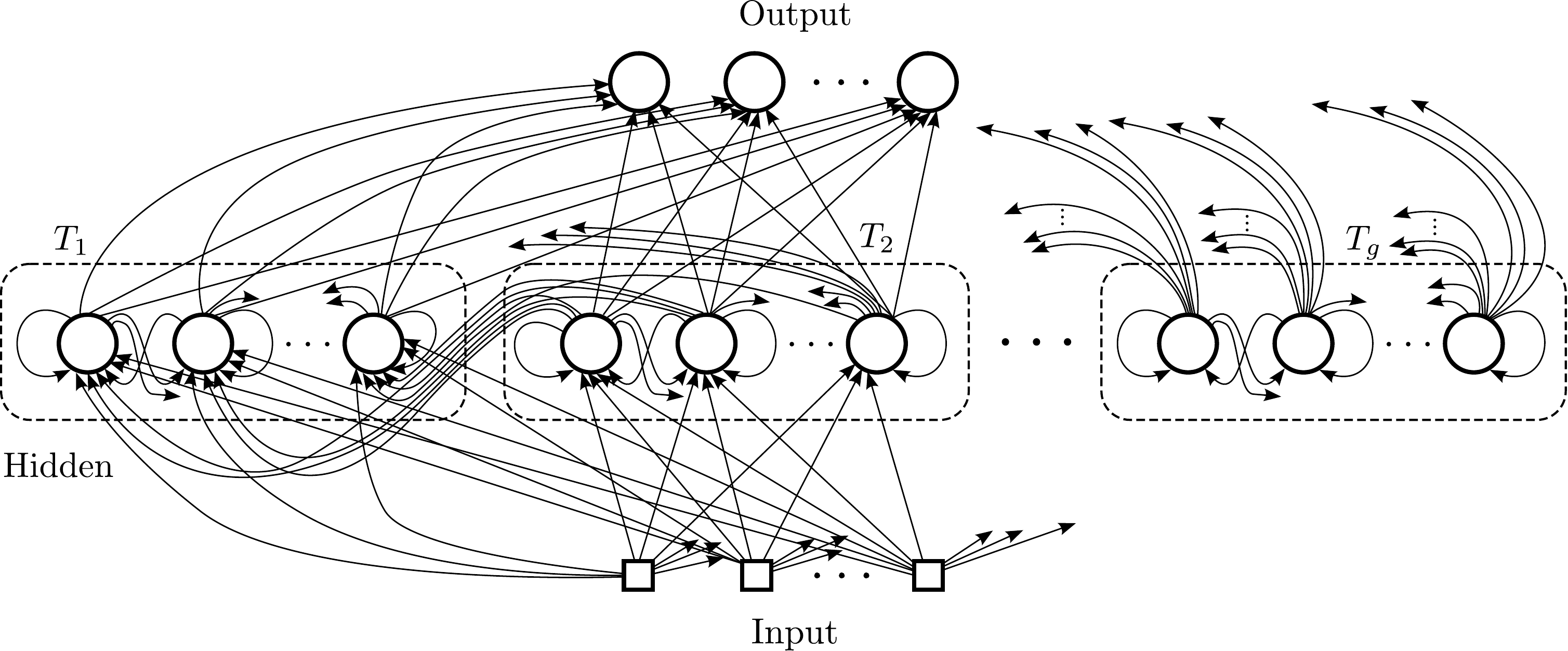}
\caption{\abr{} architecture is similar to a simple RNN with an input,
  output and hidden layer. The hidden layer is partitioned into $g$
  modules each with its own clock rate.  Within each module the
  neurons are fully interconnected. Neurons in faster module $i$ are
  connected to neurons in a slower module $j$ only if a clock period
  $T_i < T_j$.}
\label{fig:cw-rnn}
\end{figure*}

\section{Related Work}  %%%%%%%%%%%%%%%%%%%%%%%%%%%%%%%%%%%%%%%%%%%
\label{sec:related}

Contributions to the sequence modeling and recognition that are
relevant to \abr{} are introduced in this section. The primary focus
is on RNN extensions that deal with the problem of bridging long time
lags.

One model that is similar in spirit to our approach is the NARX
RNN\footnote{NARX stands for Non-linear Auto-Regressive model with
  eXogeneous inputs}~\cite{Lin:96}. But instead of simplifying the
network, it introduces an additional sets of recurrent connections
with time lags of $2$,$3$..$k$ time steps. These additional
connections help to bridge long time lags, but introduce many
additional parameters that make NARX RNN training more difficult and
run $k$ times slower.

Long Short-Term Memory (LSTM; \citealp{Hochreiter:97lstm}) uses a
specialized architecture that allows information to be stored in a
linear unit called a {\em constant error carousel} (CEC) indefinitely.
The cell containing the CEC has a set of multiplicative units (gates)
connected to other cells that regulate when new information enters the
CEC (input gate), when the activation of the CEC is output to the rest
of the network (output gate), and when the activation decays or is
"forgotten" (forget gate). These networks have been very successful
recently in speech and handwriting
recognition~\cite{graves05retraining,graves:09tpami,sak:2014arxiv}.

Stacking LSTMs into several layers
\citep{Santi:07ijcai,graves:2008nips} aims for hierarchical sequence
processing. Such a hierarchy, equipped with Connectionist Temporal
Classification (CTC;~\citealp{Graves:06icml}), performs simultaneous
segmentation and recognition of sequences.  Its deep variant currently
holds the state-of-the-art result in phoneme recognition on the TIMIT
database \cite{DBLP:conf/icassp/GravesMH13}.

Temporal Transition Hierarchy~(TTH; \citealp{Ring:93}) incrementally
adds high-order neurons in order to build a memory that is used to
disambiguate an input at the current time step. This approach can, in
principle, bridge time intervals of any length, but with
proportionally growing network size. The model was recently improved
by adding recurrent connections~\cite{DBLP:conf/aaai/Ring11} that
prevent it from bloating by reusing the high-level nodes through the
recurrent connections.

One of the earliest attempts to enable RNNs to handle long-term
dependencies is the Reduced Description
Network~\cite{Mozer:92nips,Mozer:94}. It uses leaky neurons whose
activation changes only a bit in response to its inputs. This
technique was recently picked up by Echo State
Networks (ESN; ~\citealp{jaeger:techreport2002}).

A similar technique has been used by~\citet{sutskever:10tkrnn} to
solve some serial recall tasks. These Temporal-Kernel RNNs add a
connection from each neuron to itself that has a weight that decays
exponentially in time. This is implemented in a way that can
be computed efficiently, however, its performance is still inferior to
LSTM.

Evolino~\cite{Schmidhuber:07nc,Schmidhuber:05ijcai} feeds the input to
an RNN (which can be e.g. LSTM to cope with long time lags) and then
transforms the RNN outputs to the target sequences via a optimal
linear mapping, that is computed analytically by pseudo-inverse. The
RNN is trained by an evolutionary algorithm, therefore it does not
suffer from the vanishing gradient problem. Evolino outperformed
LSTM on a set of synthetic problems and was used to
perform complex robotic manipulation \cite{mayer:iros06}.

A modern theory of why RNNs fail to learn long-term dependencies is
that simple gradient descent fails to optimize them correctly. One attempt to
mitigate this problem is Hessian Free (HF)
optimization~\cite{ICML2011Martens_532}, an adapted second-order
training method that has been demonstrated to work well with RNNs. It
allows RNNs to solve some long-term lag problems that were impossible
with stochastic gradient descent. Their performance on rather
synthetic, long-term memory benchmarks is approaching the performance of
LSTM, though the number of optimization steps in HF-RNN is
usually greater. Training networks by HF optimization is an
orthogonal approach to the network architecture, so both LSTM and
CW-RNN can still benefit from it.

HF  optimization allowed for training of Multiplicative RNN
(MRNN;~\citealp{ICML2011Sutskever_524}) that port the concept of
multiplicative gating units to SRNs. The gating units are represented
by a factored 3-way tensor in order to reduce the number of
parameters. Extensive training of an MRNN for a number of days on a
graphics cards provided impressive results in text generation tasks.

Training RNNs with Kalman filters~\cite{williams:92ekf} has shown
advantages in bridging long time lags as well, although this approach
is computationally unfeasible for larger networks.

The methods mentioned above are strictly synchronous--elements of the
network clock at the same speed. The {\em Sequence Chunker}, Neural
History Compressor or Hierarchical Temporal
  Memory~\cite{Schmidhuber:91chunker,Schmidhuber:92ncchunker}
consists of a hierarchy or stack of RNN that may run at different time
scales, but, unlike the simpler CW-RNN, it requires unsupervised event
predictors: a higher-level RNN receives an input only when the
lower-level RNN below is unable to predict it.  Hence the clock of the
higher level may speed up or slow down, depending on the current
predictability of the input stream. This contrasts the \abr{}, in
which the clocks always run at the same speed, some slower, some
faster.

\section{A Clockwork Recurrent Neural Network}  %%%%%%%%%%%%%%%%%%%%
\label{sec:method}

\name{}s (\abr{}) like SRNs, consist of input, hidden and
output layers. There are forward connections from the input to hidden
layer, and from the hidden to output layer, but, unlike the
SRN, the neurons in the hidden layer are partitioned into $g$ modules
of size $k$. Each of the modules is assigned a clock period $T_n \in
\{T_1,\dots,T_g\}$. Each module is internally fully-interconnected, but
the recurrent connections from module $j$ to module $i$ exists only if
the period $T_i$ is smaller than period $T_j$.  Sorting the modules by
increasing period, the connections between modules propagate the
hidden state  {\it right-to-left}, from slower modules
to faster modules, see \autoref{fig:cw-rnn}.

The standard RNN output, ${y}^{(t)}_O$, at a time step $t$ is calculated using the
following equations:
\begin{equation}
\mathbf{y}^{(t)}_H = f_H(\mathbf{W}_H \cdot \mathbf{y}^{(t-1)} 
                    + \mathbf{W}_I \cdot \mathbf{x}^{(t)}),
\label{eqn:hidden}
\end{equation}
\begin{equation}
\mathbf{y}^{(t)}_O = f_O(\mathbf{W}_O \cdot \mathbf{y}^{(t)}_H),
\label{eqn:output}
\end{equation}
where $\mathbf{W}_H$, $\mathbf{W}_I$ and $\mathbf{W}_O$ are the
hidden, input and output weight matrices, $\mathbf{x}_t$ is the input
vector at time step $t$, vectors $\mathbf{y}^{(t)}_H$ and
$\mathbf{y}^{(t-1)}_H$ represent the hidden neuron activations at time
steps $t$ and $t-1$.  Functions $f_H(.)$ and $f_O(.)$ are the
non-linear activation functions.  For simplicity, neuron biases are
omitted in the equations.

The main difference between \abr{} and an RNN is that at each \abr{}
time step $t$, only the output of modules $i$ that satisfy $(t$ {\sc mod} $T_i)
= 0$ are executed. The choice of the set of periods
$\{T_1,\dots,T_g\}$ is arbitrary. In this paper, we use the
exponential series of periods: module $i$ has clock period of
$T_{i}=2^{i-1}$.

\begin{figure}[t!]
\centering
\includegraphics[width=0.85\columnwidth]{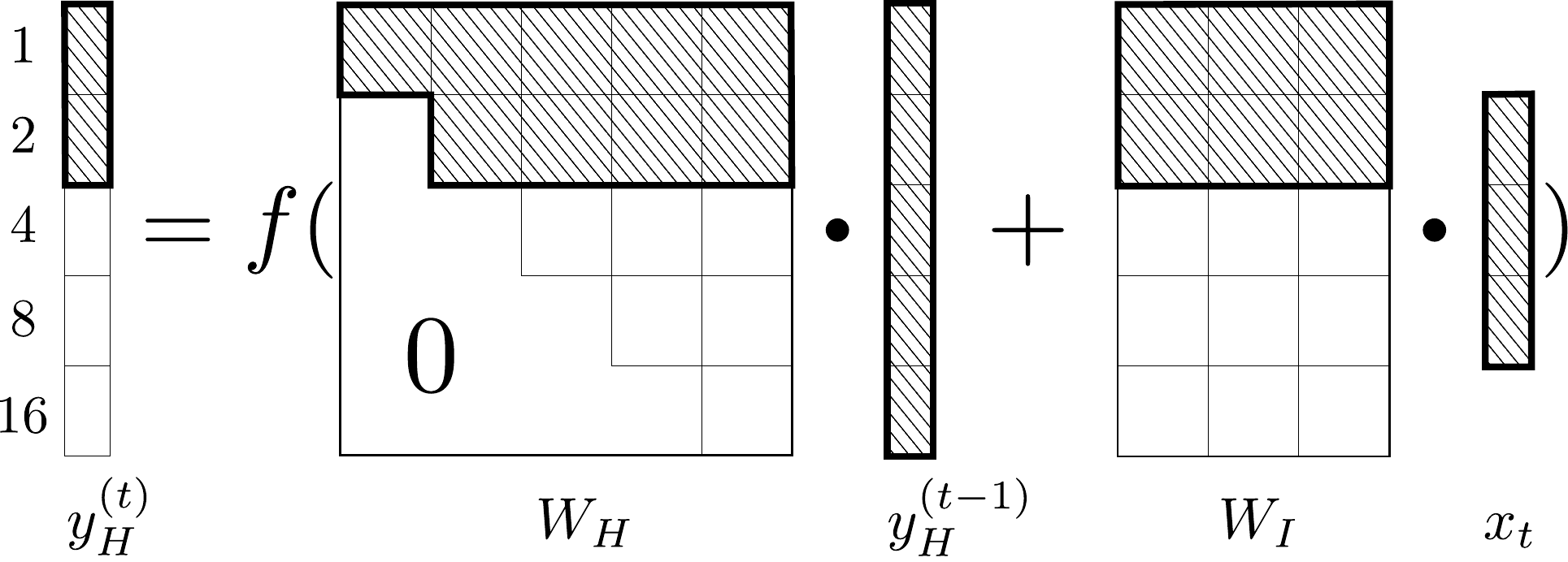}
\caption{Calculation of the hidden unit activations at time step $t=6$
  in \abr{} according to equation~\eqref{eqn:hidden}. Input and
  recurrent weight matrices are partitioned into blocks. Each block-row in
  $\mathbf{W}_H$ and $\mathbf{W}_I$ corresponds to the weights of a
  particular module. At time step $t=6$, the first two modules with
  periods $T_1=1$ and $T_2=2$ get evaluated (highlighted parts of
  $\mathbf{W}_H$ and $\mathbf{W}_I$ are used) and the highlighted
  outputs are updated. Note that, while using exponential series of
  periods, the active parts of $\mathbf{W}_H$ and $\mathbf{W}_I$ are
  always contiguous.}
\label{fig:formula}
\end{figure}

Matrices $\mathbf{W}_H$ and $\mathbf{W}_I$ are partitioned into $g$
blocks-rows:
\begin{equation}
\mathbf{W}_H = 
\begin{pmatrix}
\mathbf{W}_{H_1}\\
\vdots\\
\mathbf{W}_{H_g}
\end{pmatrix}
\qquad
\mathbf{W}_I = 
\begin{pmatrix}
\mathbf{W}_{I_1}\\
\vdots\\
\mathbf{W}_{I_g}
\end{pmatrix}
\end{equation}
and $\mathbf{W}_H$ is a block-upper triangular matrix, where each block-row,
$\mathbf{W}_{H_i}$, is partitioned into block-columns
$\{\mathbf{0}_1,\dots,\mathbf{0}_{i-1},\mathbf{W_{H_{i,i}},\dots,\mathbf{W_{H_{i,g}}}
}\}$. At each forward pass time step, only the block-rows of $\mathbf{W}_H$ and
$\mathbf{W}_I$ that correspond to the executed modules are used for
evaluation in \autoref{eqn:hidden}:
\begin{equation}
\mathbf{W}_{H_i} = \left\{ 
\begin{array}{cl}
\mathbf{W}_{H_i} & \textrm{for } (t \textrm{~{\sc mod}~} T_{i}) = 0 \\
\mathbf{0} & \textrm{otherwise}
\end{array}\right.
\end{equation}
and the corresponding parts of the output vector, $\mathbf{y}_H$, are
updated. The other modules retain their output values from the
previous time-step.  Calculation of the hidden activation at time step
$t=6$ is illustrated in \autoref{fig:formula}.

\begin{figure}[t]
\vskip 0.2in
\begin{center}
\centerline{\includegraphics[width=\columnwidth]{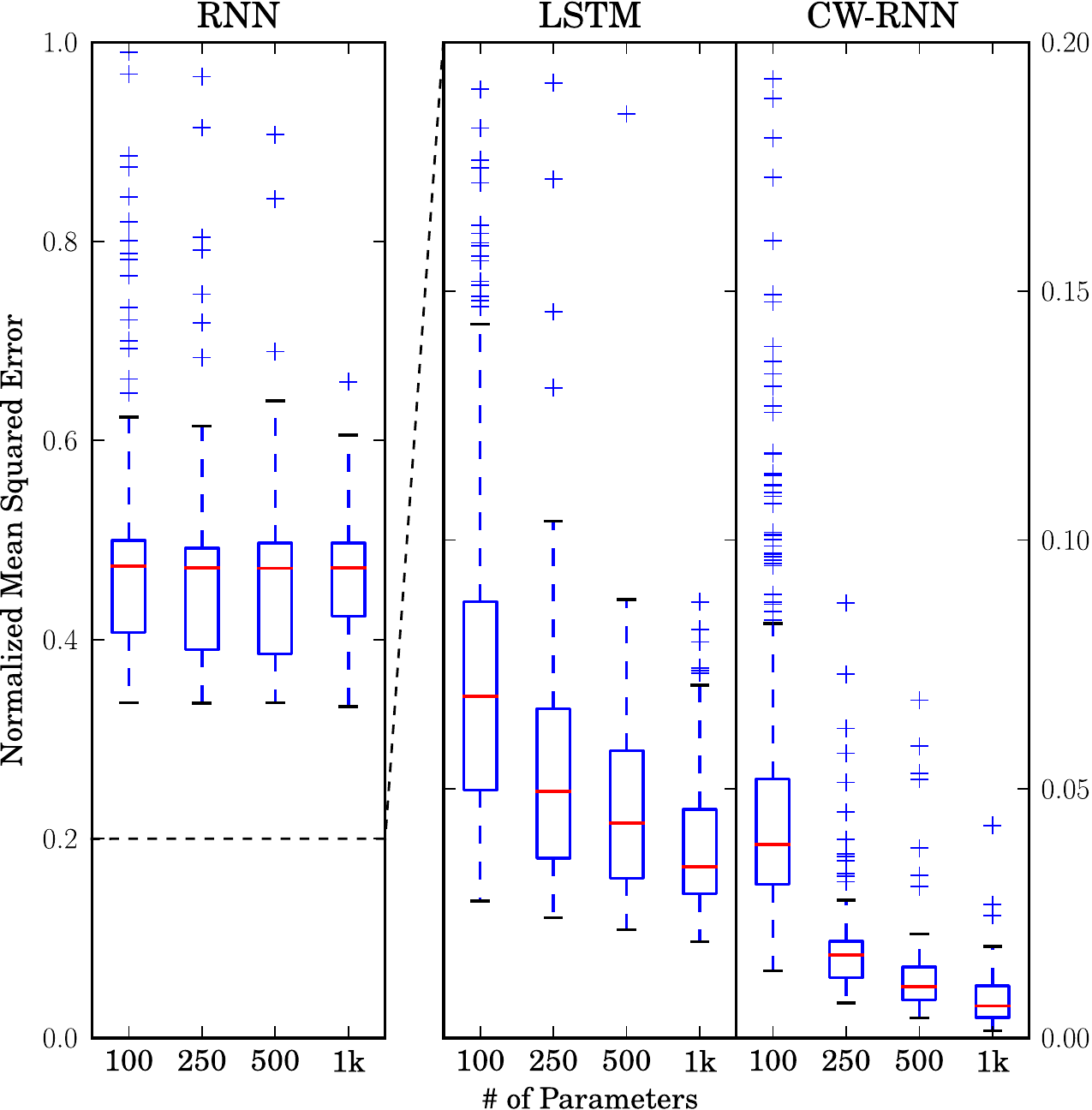}}
\caption{ The normalized mean squared error for the sequence
  generation task, divided into one column per method, with one
  box-whisker (showing mean value, 25\% and 75\% quantiles, minimum,
  maximum and outliers) for every tested size of the network.  Note
  that the plot for the standard RNN has a different scale than the
  other two.}
\label{fig:results-sg}
\end{center}
\vskip -0.2in
\end{figure} 

\begin{figure}[ht]
\vskip 0.2in
\begin{center}
\centerline{\includegraphics[width=\columnwidth]{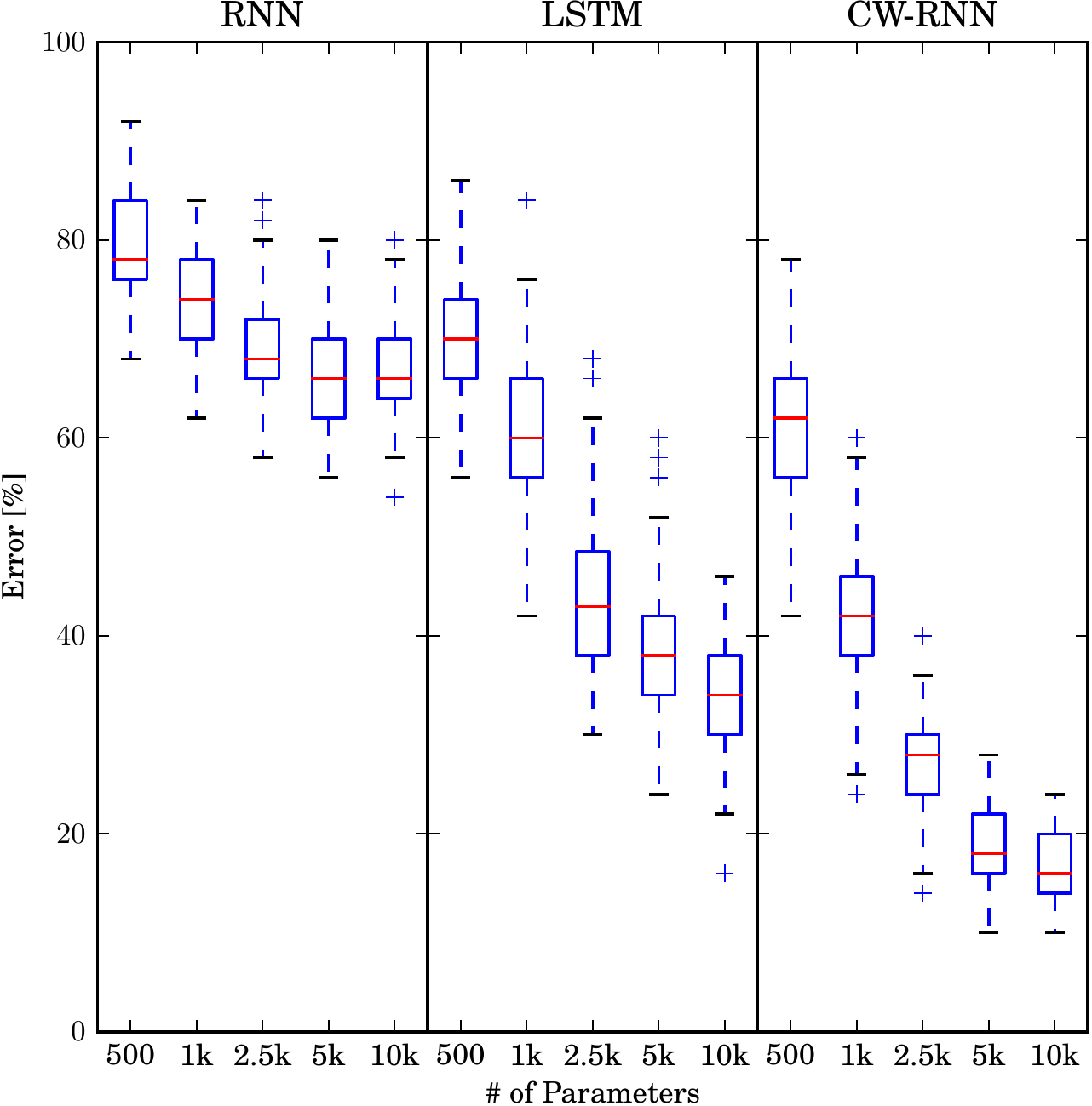}}
\caption{ The classification error for the word classification task,
  divided into three columns (one per method), with one box-whisker
  for every tested network size.  }
\label{fig:results-wc}
\end{center}
\vskip -0.2in
\end{figure}

As a result, the low-clock-rate modules process, retain and output the
long-term information obtained from the input sequences (not being
distracted by the high speed modules), whereas the high-speed modules
focus on the local, high-frequency information (having the context
provided by the low speed modules available).

The backward pass of the error propagation is similar to RNN as
well. The only difference is that the error propagates only from
modules that were executed at time step $t$. The error of
non-activated modules gets copied back in time (similarly to copying
the activations of nodes not activated at the time step $t$ during the
corresponding forward pass), where it is added to the back-propagated
error.

\abr{} runs much faster than a simple RNN with the same number of
hidden nodes since not all modules are evaluated at every time
step. The lower bound for the \abr{} speedup compared to an RNN with
the same number of neurons is $g/4$ in the case of this exponential
clock setup, see Appendix for a detailed derivation.

\section{Experiments}  %%%%%%%%%%%%%%%%%%%%%%%%%%%%%%%%%%%%%%%%%%%%
\label{sec:exp}

\abr s were compared to the simple RNN (SRN) and LSTM networks. All
networks have one hidden layer with the \emph{tanh} activation
function, and the number of nodes in the hidden layer was chosen to
obtain (approximately) the same number of parameters for all three
methods (in the case of \abr, the clock periods were included in the
parameter count).

Initial values for all the weights were drawn from a Gaussian
distribution with zero mean and standard deviation of $0.1$. Initial
values of all internal state variables were set to $0$. Each setup was
run $100$ times with different random initialization of
parameters. All networks were trained using Stochastic Gradient
Descent (SGD) with Nesterov-style
momentum~\cite{SutskeverMartensDahlHinton_icml2013}.

\subsection{Sequence Generation}
\label{sec:exp:gen}

\begin{figure*}[ht]
\vskip 0.2in
\begin{center}
\centerline{\includegraphics[width=\textwidth]{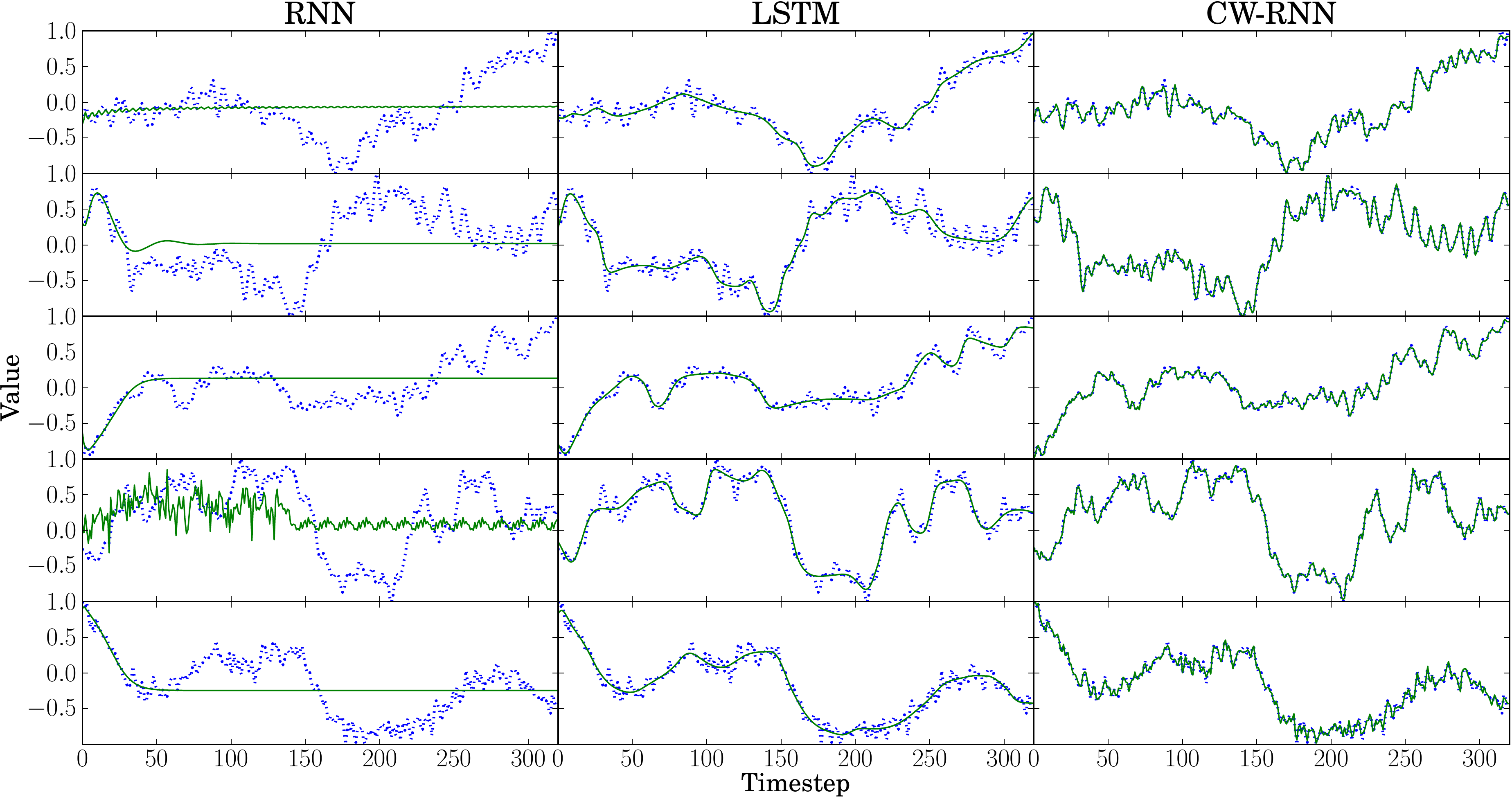}}
\caption{Output of the best performing network (solid, green) compared
  to the target signal (dotted, blue) for each method (column) and
  each training sequence (row). RNN tends to learn the first few steps of
  the sequence and then generates the mean of the remaining portion,
  while the output of LSTM resembles a sliding average, and {\abr}
  approximates the sequence much more accurately.}
\label{fig:outputs-sg}
\end{center}
\vskip -0.2in
\end{figure*} 

The goal of this task is to train a recurrent neural network, that
receives no input, to generate a target sequence as accurately as
possible.  The weights of the network can be seen as a (lossy)
encoding of the whole sequence, which could be used for compression.

Five different target sequences were created by sampling a piece of
music\footnote{taken from the beginning of the first track
  {\it Man\'{y}rista} of album {\it Musica Deposita} by {\it Cuprum}}
at $44.1$\,Hz for $7$\,ms. The resulting sequences of $320$ data points
each were scaled to the interval $[-1,1]$. In the following
experiments we compare performance on these five sequences.

All networks used the same architecture: no inputs, one hidden layer
and a single linear output neuron.  Each network type was run with 4
different sizes: $100$, $250$, $500$, and $1000$ parameters, see
\autoref{tab:gen} for the summary of number of hidden nodes.  The
networks were trained over $2000$ epochs to minimize the mean squared
error. After that time the error no longer decreased
noticeably. Momentum was set to $0.95$ while the learning rate was
optimized separately for every method, but kept the same for all
network sizes.

A learning rate of $3\times 10^{-4}$ was found to be optimal for RNN
and \abr{} while for LSTM $3\times 10^{-5}$ gave better results.  For
LSTM it was also crucial to initialize the bias of the forget gates to
a high value (5 in this case) to encourage the long-term memory. The
hidden units of \abr{} were divided into nine equally sized groups
with exponential clock-timings \{$1$, $2$, $4$, \dots, $256$\}.

The results for the experiments are shown in
\autoref{fig:results-sg}.  It is obvious that RNNs fail to generate
the target sequence, and they do not seem to improve with network
size. LSTM does much better, and shows an improvement as the networks
get bigger. {}\abr s give by far the best results, with the smallest
one being roughly on par with the second-biggest LSTM network. Also,
all but the smallest \abr{} have significantly less variance than all
the other methods. To get an intuitive understanding of what is
happening, \autoref{fig:outputs-sg} shows the output of the best
network of each type on each one of the five audio samples. The average error of the
best networks is summarized in \autoref{tab:summary} (row 1).

\begin{table}
\centering
\caption{Number of hidden neurons (cells in the case 
of LSTM) for RNN, LSTM and \abr{} for each network size specified in terms of
the number of parameters (weights) for the sequence generation task.}
\vspace{1.5mm}
\begin{tabular}{cccc}
\toprule
{\bf \# of Parameters} & {\bf RNN} & {\bf LSTM} & {\bf CW-RNN}\tabularnewline
\midrule
100 & 9 & 4 & 11\tabularnewline
250 & 15 & 7 & 19\tabularnewline
500 & 22 & 10 & 27\tabularnewline
1 000 & 31 & 15 & 40\tabularnewline
\bottomrule
\end{tabular}
\label{tab:gen}
\end{table}

\subsection{Spoken Word Classification}
\label{sec:exp:wc}

The second task is sequence classification instead of generation.
Each sequence contains an audio signal of one spoken word from the
TIMIT Speech Recognition Benchmark~\cite{timit}. The dataset contains
$25$ different words (classes) arranged in $5$ clusters based on their
suffix. Because of the suffix-similarity the network needs to learn
long-term dependencies in order to disambiguate the words. The words
are:
\begin{description}
\setlength{\itemsep}{0em}
\item[{Cluster 1:}] {\tt \footnotesize making, walking, cooking, looking, working} 
\item[{Cluster 2:}]  {\tt \small biblical, cyclical, technical, classical, critical}
\item[{Cluster 3:}]  {\tt \small tradition, addition, audition, recognition, competition} 
\item[{Cluster 4:}] {\tt \small musicians, discussions, regulations, accusations, conditions} 
\item[{Cluster 5:}]  {\tt \small subway, leeway, freeway, highway, hallway}
\end{description}
For every word there are $7$ examples from different speakers, which
were partitioned into $5$ for training and $2$ for testing, for a
total of 175 sequences ($125$ train, $50$ test). Each sequence element
consists of $12$-dimensional MFCC vector~\cite{Mermelstein76} plus
energy, sampled every $10$\,ms over a $25$\,ms window with a
pre-emphasis coefficient of $0.97$. Each of the $13$ channels was then
normalized to have zero mean and unit variance over the whole training
set.

All network types used the same architecture: $13$ inputs, a single
hidden and a softmax output layer with $25$ units. Five hidden layer
sizes were chosen such that the total number of parameters for the
whole network is roughly $0.5$k, $1$k, $2.5$k, $5$k, and $10$k.

All networks used a learning rate of $3\times 10^{-4}$, a momentum of 0.9,
and were trained to minimize the Multinomial Cross Entropy Error. 
Every experiment was repeated $100$ times with different
random initializations. 

Because the dataset is so small, Gaussian noise with a standard
deviation of $0.6$ was added to the inputs during training to guard against overfitting.
Training was stopped once the error on the \emph{noise-free} training
set did not decrease for $5$ epochs. To obtain good results with LSTM,
it was again important to initialize the forget gate bias to $5$. For the
\abr{} the neurons were divided evenly into $7$ groups with exponentially
increasing periods: \{$1$, $2$, $4$, $8$, $16$, $32$, $64$\}.

\autoref{fig:results-wc} shows the classification error of the
different networks on the word classification task. Here again, RNNs
perform the worst, followed by LSTMs, which give substantially better
results, especially with more parameters. {}\abr s beat both RNN and
LSTM networks by a considerable margin of $8$-$20$\% on average
irrespective of the number of parameters. The error of the largest
networks is summarized in \autoref{tab:summary} (row 2).

\begin{table}
\centering
\caption{Number of hidden neurons (cells in the case 
of LSTM) for RNN, LSTM and \abr{} for each network size specified in terms of
the number of parameters (weights) for the spoken word classification task.}
\vspace{1.5mm}
\begin{tabular}{cccc}
\toprule
{\bf \# of Parameters} & {\bf RNN} & {\bf LSTM} & {\bf CW-RNN}\tabularnewline
\midrule
500 & 10 & 5 & 10\tabularnewline
1000 & 18 & 8 & 19\tabularnewline
2500 & 34 & 17 & 40\tabularnewline
5000 & 54 & 26 & 65\tabularnewline
10000 & 84 & 41 & 102\tabularnewline
\bottomrule
\end{tabular}
\label{tab:word}
\end{table}

\begin{table}
\centering
\caption{Mean error and standard deviation (averaged over $100$
  runs) for the largest (best) LSTM, RNN and CW-RNN on both tasks.
  \abr{} is $5.7\times$ better than LSTM on Task~\ref{sec:exp:gen},
  sequence generation, and more than $2\times$ better than LSTM on
  Task~\ref{sec:exp:wc}, spoken word classification.}
\vspace{1.5mm}
\begin{tabular}{l@{\hskip 1.6mm}r@{$\pm$}lr@{$\pm$}lr@{$\pm$}l}
\toprule
{\bf Task} & \multicolumn{2}{c}{{\bf RNN}} & 
             \multicolumn{2}{c}{{\bf LSTM}} & 
             \multicolumn{2}{c}{{\bf CW-RNN}}\tabularnewline
\midrule
\ref{sec:exp:gen}~ NMSE & 0.46 & 0.08 & 0.04 & 0.01 & 0.007 & 0.004 \tabularnewline
\ref{sec:exp:wc}~ Error [\%]& 66.8   &  4.7   & 34.2   & 5.6    & 16.8   & 3.5    \tabularnewline
\bottomrule
\end{tabular}
\label{tab:summary}
\end{table}

\section{Discussion} %%%%%%%%%%%%%%%%%%%%%%%%%%%%%%%%%%%%%%%%%%%%%%
\label{sec:disc}
The experimental results show that the simple mechanism of running
subsets of neurons at different speeds allows an RNN to efficiently
learn the different dynamic time-scales inherent in complex signals.

Other functions could be used to set the module periods: linear,
Fibonacci, logarithmic series, or even fixed random periods. These
were not considered in this paper because the intuitive setup of using
an exponential series worked well in these preliminary experiments.
Another option would be to learn the periods as well, which, to use
error back-propagation would require a differentiable modulo function
for triggering the clocks. Alternatively, one could train the clocks
(together with the weights) using evolutionary algorithms which do not
require a closed form for the gradient. Note that the lowest period in
the network can be greater than $1$. Such a network would not be able
to change its output at every time step, which may be useful as a
low-pass filter when the data contains noise.

Also, the modules do not have to be all of the same size. One could
adjust them according to the expected information in the input
sequences, by e.g. using frequency analysis of the data and setting up
modules sizes and clocks proportional to the spectrum.

Grouping hidden neurons into modules is a partway to having each
weight have its own clock. Initial experiments, not included in this
paper, have shown that such networks are hard to train and do not
provide good results.

\abr{} showed superior performance on the speech 
data classification among all three models tested. Note that, unlike
in the standard approach, in which the speech signal frequency
coefficients are first translated to phonemes which are modeled
with a standard approach like Hidden Markov Modes for complete
words, \abr{} attempts to model and recognize the complete words
directly, where it benefits from the modules running at multiple speeds.

Future work will start by conducting a detailed analysis of the
internal dynamics taking place in the \abr{} to understand how the
network is allocating resources for a given type of input sequence.
Further testing on other classes of problems, such as reinforcement
learning, and comparison to the larger set of connectionist models for
sequential data processing are also planned.

\section*{Appendix} %%%%%%%%%%%%%%%%%%%%%%%%%%%%%%%%%%%%%%%%%%%%%%%
\label{sec:appendix}
\abr{} has fewer total parameters and even fewer operations
per time step than a standard RNN with the same number of neurons.
Assume \abr{} consists of $g$ modules of size $k$ for a total
of $n=kg$ neurons. Because a neuron is only connected to other neurons 
with the same or larger period, the number of parameters $N_{H}$ for
the recurrent matrix is:
\[
N_{H}=\sum_{i=1}^{g}\sum_{j=1}^{k}k (g-i+1)=k^{2}\sum_{i=0}^{g-1}(g-i)=\frac{n^{2}}{2}+\frac{nk}{2}.
\]
Compared to the $n^{2}$ parameters in the recurrent matrix
$\mathbf{W}_H$ of RNN this results in roughly half as many parameters:
\[
\frac{N_{H}}{n^{2}}=\frac{\frac{n^{2}}{2}+\frac{nk}{2}}{n^{2}}=\frac{n^{2}+nk}{2n^{2}}=\frac{n+k}{2n}=\frac{g+1}{2g}\approx\frac{1}{2}.
\]
Each module $i$ is evaluated only every $T_{i}$-th time step,
therefore the number of operations at a time step is:
\[
O_{H}=k^{2}\sum_{i=0}^{g-1}\frac{g-i}{T_{i}}.
\]
For exponentially scaled periods, $T_{i}=2^{i}$, the upper bound for
number of operations, $O_H$, needed for $\mathbf{W}_H$ per time step
is:
\begin{multline*}
O_{h}=k^{2}\sum_{i=0}^{g-1}\frac{g-i}{2^{i}}=k^{2}\Biggl(g\underbrace{\sum_{i=0}^{g-1}\frac{1}{2^{i}}}_{\leq2}+\underbrace{\sum_{i=0}^{g-1}\frac{i}{2^{i}}}_{\leq2}\Biggr)\leq\\ 
\leq k^{2}(2g-2)\leq2nk,
\end{multline*}
because $g\geq2$ this is less than or equal to $n^{2}$. 
Recurrent operations in \abr{} are faster than in an RNN with
the same number of neurons by a factor of at least $g/2$,
which, for typical \abr{} sizes ends up being between 2 and 5. 
Similarly, upper bound for the number of input weight
evaluations, $E_{I}$, is:
\begin{multline*}
O_{I}=\sum_{i=0}^{g-1}\frac{km}{T_{i}}=km\sum_{i=0}^{g-1}\frac{1}{T_{i}}\leq2km
\end{multline*}
Therefore, the overall \abr{} speed-up w.r.t RNN is:
\begin{multline*}
\frac{n^{2}+nm+n}{O_{R}+O_{I}+2n}=\frac{k^{2}g^{2}+kgm+kg}{k^{2}(2g-2)+2km+2kg}=\\
=\frac{g(kg+m+1)}{2(k(g-1)+m+g)}=\frac{g}{2}\underbrace{\frac{(kg+m+1)}{k(g-1)+m+g}}_{\geq\frac{1}{2}}\geq\frac{g}{4}
\end{multline*}
Note that this is a conservative lower bound.

\section*{Acknowledgments}
This research was supported by Swiss National Science Foundation grant
\#138219: ``Theory and Practice of Reinforcement Learning 2'', and the
EU FP7 project ``NanoBioTouch'', grant \#228844.

\bibliography{bib}  %%%%%%%%%%%%%%%%%%%%%%%%%%%%%%%%%%%%%%%%%%%

\bibliographystyle{icml2014}

\end{document}